\documentclass[runningheads]{llncs}
\usepackage{graphicx}

\usepackage{tikz}
\usepackage{comment}
\usepackage{amsmath,amssymb} 
\usepackage{color}

\usepackage{soul}
\usepackage{hyperref}
\begin{document}
\pagestyle{headings}
\mainmatter
\def\ECCVSubNumber{5494}  

\title{Tracking Emerges by Looking Around Static Scenes, with Neural 3D Mapping}

\titlerunning{Tracking Emerges with Neural 3D Mapping}
%
\author{Adam W. Harley \and Shrinidhi K. Lakshmikanth \and \\Paul Schydlo \and Katerina Fragkiadaki}
\authorrunning{A. W. Harley et al.}
%
\institute{Carnegie Mellon University, Pittsburgh PA, 15213, USA \\
\email{\{aharley,kowshika,pschydlo,katef\}@cs.cmu.edu}}
\maketitle

\newcommand\adam[1]{\textcolor{magenta}{#1}}
\newcommand\todo[1]{\textcolor{red}{#1}}
\newcommand\red[1]{\textcolor{red}{#1}}
\newcommand\gist[1]{\textcolor{cyan}{#1}}

\newcommand{\M}{\mathcal{M}} 
\newcommand{\R}{\mathcal{R}} 
\newcommand{\Mt}{\mathcal{M}^{(t)}} 
\newcommand{\Mz}{\mathcal{M}^{(0)}} 
\newcommand{\Mi}{\mathcal{M}^{(i)}} 
\newcommand{\Mj}{\mathcal{M}^{(j)}} 
\newcommand{\Mto}{\mathcal{M}^{(t+1)}} 
\newcommand{\I}{I} 
\newcommand{\D}{D} 

\newcommand{\loss}{\mathcal{L}}

\begin{abstract}
We hypothesize that an agent that can look around in static scenes can learn rich visual representations applicable to 3D object tracking in complex dynamic scenes. We are motivated in this pursuit by the fact that the physical world itself is mostly static, and multiview correspondence labels are relatively cheap to collect in static scenes, e.g., by triangulation. We propose to leverage multiview data of \textit{static points} in arbitrary scenes (static or dynamic), to learn a neural 3D mapping module which produces features that are correspondable across time. The neural 3D mapper consumes RGB-D data as input, and produces a 3D voxel grid of deep features as output. We train the voxel features to be correspondable across viewpoints, using a contrastive loss, and correspondability across time emerges automatically. At test time, given an RGB-D video with approximate camera poses, and given the 3D box of an object to track, we track the target object by generating a map of each timestep and locating the object's features within each map. In contrast to models that represent video streams in 2D or 2.5D, our model's 3D scene representation is disentangled from projection artifacts, is stable under camera motion, and is robust to partial occlusions. We test the proposed architectures in challenging simulated and real data, and show that our unsupervised 3D object trackers outperform prior unsupervised 2D and 2.5D trackers, and approach the accuracy of supervised trackers. This work demonstrates that 3D object trackers can emerge without tracking labels, through multiview self-supervision on static data. 

\end{abstract}

\section{Introduction} \label{sec:intro}


A large part of the real world almost never moves. 
This may be surprising, since moving entities easily attract our attention \cite{franconeri2003moving}, and because we ourselves spend most of our waking hours continuously in motion. Objects like roads and buildings, however, stay put. 
Can we leverage this property of the world to learn visual features suitable for interpreting complex scenes with many moving objects? 

In this paper, we hypothesize that a correspondence module learned in static scenes will also work well in dynamic scenes. This is motivated by the fact that the content of dynamic scenes is the same as the content of static scenes. We would like our hypothesis to be true, because correspondences are far cheaper to obtain in static scenes than in dynamic ones. In static scenes, one can simply deploy a Simultaneous Localization and Mapping (SLAM) module to obtain a 3D reconstruction of the scene, and then project reconstructed points back into the input imagery to obtain multiview correspondence labels. Obtaining correspondences in dynamic scenes would require taking into account the motion of the objects (i.e., tracking). 

We propose to leverage multiview data of static points in arbitrary scenes (static or dynamic), 
to learn a neural 3D mapping module which produces features that are correspondable across viewpoints \textit{and} timesteps. The neural 3D mapper consumes RGB-D (color and depth) data as input, and produces a 3D voxel grid of deep features as output. We train the voxel features to be correspondable across viewpoints, using a contrastive loss. At test time, given an RGB-D video with approximate camera poses, and given the 3D box of an object to track, we track the target object by generating a map of each timestep and locating the object's features within each map. 



In contrast to models that represent video streams in 2D \cite{DBLP:journals/corr/WangG15a,DBLP:journals/corr/abs-1903-07593,vondrick2018tracking}, our model's 3D scene representation is disentangled from camera motion and projection artifacts. 
This provides an inductive bias that scene elements maintain their size and shape across changes in camera viewpoint, and reduces the need for scale invariance in the model's parameters. Additionally, the stability of the 3D map under camera motion allows the model to constrain correspondence searches to the 3D area where the target was last seen, which is a far more reliable cue than 2D pixel coordinates. In contrast to models that use 2.5D representations (e.g., scene flow \cite{Menze2015CVPR}), our neural 3D maps additionally provide features for partially occluded areas of the scene, since the model can infer their features from context. This provides an abundance of additional 3D correspondence candidates at test time, as opposed to being limited to the points observed by a depth sensor. 

Our work builds on geometry-aware recurrent neural networks (GRNNs) \cite{commonsense}. GRNNs are modular differentiable neural architectures that take as input RGB-D streams of a static scene under a moving camera and infer 3D feature maps of the scene, estimating and stabilizing against camera motion. The work of Harley et al.~\cite{adam3d} showed that training GRNNs for contrastive view prediction in static scenes helps semi-supervised 3D object detection, as well as moving object discovery. In this paper, we extend those works to learn from dynamic scenes with independently moving objects, and simplify the GRNN model by reducing the number of losses and modules. 
Our work also builds on the work of Vondrick et al.~\cite{vondrick2018tracking}, which showed that 2D pixel trackers can emerge without any tracking labels, through self-supervision on a colorization task. In this work, we show that 3D voxel trackers can emerge without tracking labels, through contrastive self-supervision on static data. In the fact that we learn features from correspondences established through triangulation, our work is similar to Dense Object Nets~\cite{florence2018dense}, though that work used object-centric data, and used background subtraction to apply loss on static objects, whereas in our work we do moving-object subtraction to apply loss on \textit{anything} static. We do not assume a priori that we know which objects to focus on. 


We test the proposed architectures in simulated and real datasets of urban driving scenes (CARLA~\cite{Dosovitskiy17} and KITTI~\cite{Geiger2013IJRR}). We evaluate the learned 3D visual feature representations on their ability to track objects over time in 3D. We show that the learned visual feature representations can accurately track objects in 3D, simply supervised by observing static data. Furthermore, our method outperforms 2D and 2.5D baselines, demonstrating the utility of learning a 3D representation for this task instead of a 2D one. 

The main contribution of this paper is to show that learning feature correspondences from static 3D points causes 3D object tracking to emerge. We additionally introduce a neural 3D mapping module which simplifies prior works on 3D inverse graphics, and learns from a simpler objective than considered in prior works. 
Our code and data are publicly available\footnote{\url{https://github.com/aharley/neural_3d_tracking}}.

\section{Related  Work} \label{sec:related}

\subsection{Learning to see by moving}

Both cognitive psychology and computational vision have realised the importance of motion for the development of visual perception \cite{Gibson1979-GIBTEA,wiskott2002slow}.  
Predictive coding theories \cite{rao1999predictive,friston2003learning} suggest that the brain
predicts observations at various levels of abstraction; temporal prediction is thus of central interest. These theories currently have extensive empirical support: stimuli are processed more quickly if they are predictable \cite{mcclelland1981interactive,pinto2015expectations}, prediction error is reflected in increased neural activity \cite{rao1999predictive,brodski2015faces}, and disproven expectations lead to learning \cite{schultz1997neural}. Several computational models of frame predictions have been proposed   \cite{rao1999predictive,Eslami1204,DBLP:journals/corr/TatarchenkoDB15,commonsense,oord2018representation}. 
Alongside future frame prediction, predicting some form of contextual or missing information has also been explored, such as 
predicting frame ordering \cite{lee2017unsupervised}, temporal instance-level associations \cite{DBLP:journals/corr/WangG15a}
color from grayscale \cite{vondrick2018tracking}, egomotion  \cite{DBLP:journals/corr/JayaramanG15,DBLP:journals/corr/AgrawalCM15} and motion trajectory forecasting \cite{DBLP:journals/corr/WalkerDGH16}. Most of these unsupervised methods are evaluated as pre-training mechanisms for object detection or classification \cite{DBLP:journals/corr/MisraZH16,DBLP:journals/corr/WalkerDGH16,DBLP:journals/corr/WangG15a,adam3d}. 

Video motion segmentation literature  explores the use of videos in unsupervised moving object discovery \cite{Pont-Tuset_arXiv_2017}.  
Most motion segmentation methods operate in 2D image space, and cluster 2D optical flow vectors or 2D flow trajectories  to segment moving objects \cite{OB11,springerlink:10.1007/978-3-642-15555-0_21,Fragkiadaki:topology}, or use low-rank trajectory constraints \cite{DBLP:conf/iccv/CheriyadatR09,Tomasi:1992:SMI:144398.144403,10.1109/ICCV.1995.466815}. 
Our work differs in  that we address object detection and segmentation in 3D as opposed to 2D, by estimating 3D motion of the ``imagined" (complete) scene, as opposed to 2D motion of the  pixel observation stream.


\subsection{Vision as inverse graphics}

Earlier works in Computer Vision proposed  casting visual recognition as  inverse rendering \cite{bruno,yuillek06}, as opposed to feedforward labelling. The ``blocks world" of Roberts \cite{blocksworld} had the goal of reconstructing the 3D scene depicted in the image in terms of 3D solids found in a database.  
A key question to be addressed is: what representations should we use for the  intermediate latent 3D structures? Most works seek to map images to explicit 3D representations, such as 3D pointclouds \cite{Wu2016,DBLP:journals/corr/TungHSF17,tinghuisfm,sfmnet}, 3D meshes \cite{Loper:2015:SSM:2816795.2818013,DBLP:journals/corr/abs-1711-07566}, or binary 3D voxel occupancies \cite{DBLP:journals/corr/TulsianiZEM17,LSM,3Dshapenets}.
The aforementioned manually designed 3D representations, e.g., 3D meshes, 3D keypoints, 3D pointclouds, 3D voxel grids, may not be general enough to express the rich 3D world, which contains liquids, deformable objects, clutter, dirt, wind, etc., and at the same time may be over descriptive when detail is unnecessary. In this work, we opt for \textit{learning-based 3D feature representations} extracted end-to-end from the RGB-D input stream as proposed by Tung et al.~\cite{commonsense} and Harley et al.~\cite{adam3d}. We extend the architectures of those works to handle and learn from videos of dynamic scenes, as opposed to only videos of static scenes.  



\begin{figure}[t!]
 \centering
 \includegraphics[width=1.0\linewidth]{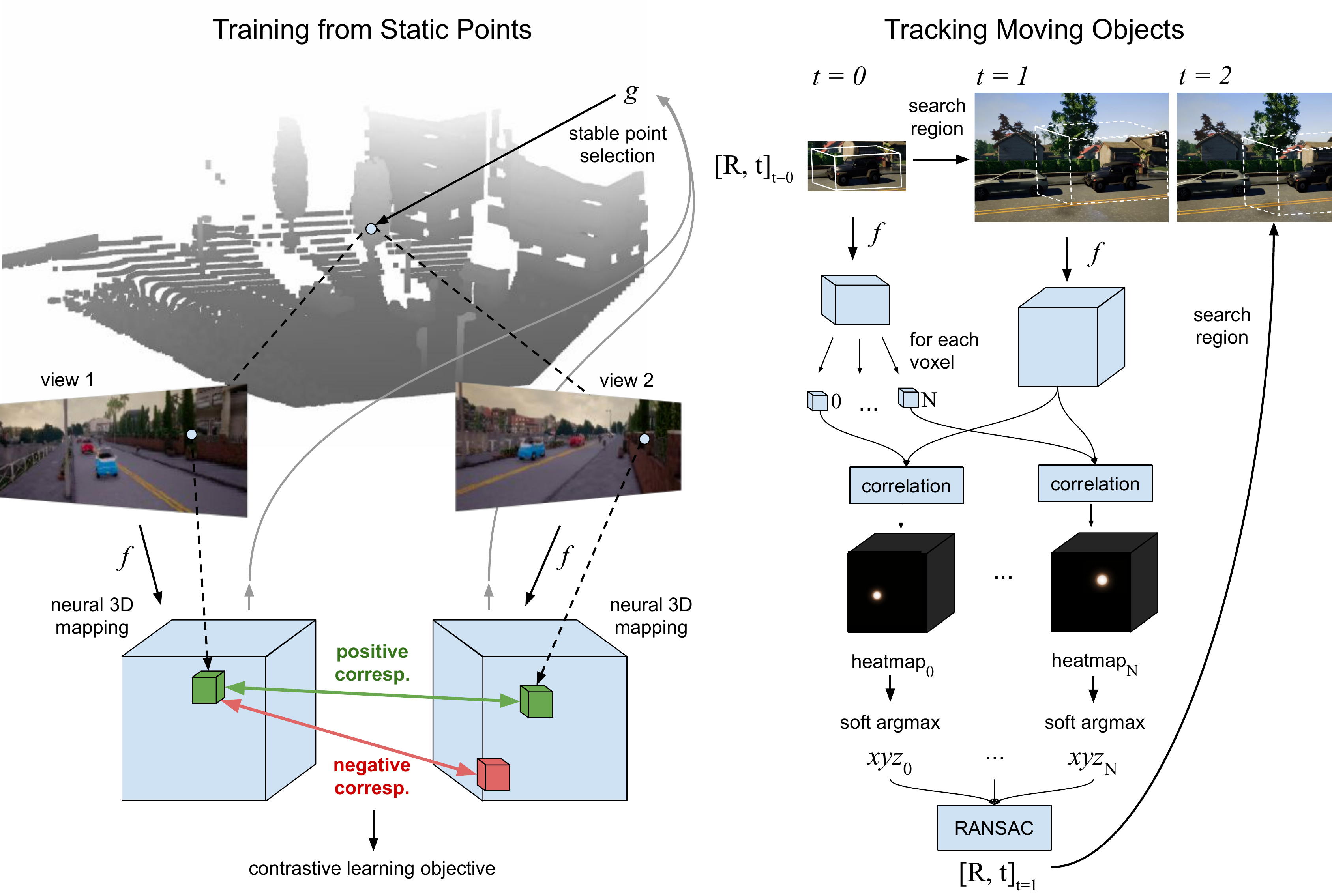}
 \caption{ 
 \textbf{Tracking emerges by looking around in static scenes, with neural 3D mapping.} \textit{Left: Training regime.} Our learned 3D neural mapper $f$ maps RGB-D inputs into featurized 3D scene maps. Points that correspond across multiple views, provided by static data or selected by a reliability network $g$, provide labels for a contrastive learning objective. \textit{Right: Testing regime.} Given an object box in the zeroth frame, our encoder creates features for the object, and searches for re-occurences of those features in the 3D scene maps of future time steps. The model then explains the full-object motion via a robust estimate of the rigid transformation across frames. Each estimated transformation is used to initialize the search region in the next timestep. Our model does not use any human supervision, and never trains explicitly for tracking; instead, it is supervised only by observing static points in 3D scenes. 
 }
 \label{fig:overview}
\end{figure}

\section{Learning to Track with Neural 3D Mapping} \label{sec:model}

We consider a mobile agent that can move around in a 3D scene, and observe it from multiple viewpoints. The scene can optionally contain dynamic (moving and/or deforming) objects. The agent has an RGB camera with known intrinsics, and a depth sensor registered to the camera's coordinate frame. 
It is reasonable to assume that a mobile agent who moves at will has access to its approximate egomotion, since it chooses where to move and what to look at \cite{patla1991visual}. In simulation, we use ground truth camera poses; in real data, we use approximate camera poses provided by an inertial navigation system (GPS and IMU). In simulation, we use random viewpoints; in real data, we use just one forward-facing camera (which is all that is available). Note that a more sophisticated mobile agent might attempt to select viewpoints intelligently at training time. 

Given the RGB-D and egomotion data, our goal is to learn 3D feature representations that can correspond entities across time, despite variations in pose and appearance. We achieve this by training inverse graphics neural architectures that consume RGB-D videos and infer 3D feature maps of the full scenes, as we describe in Section~\ref{sec:neural_mapping}. To make use of data where some parts are static and other parts are moving, we learn to identify static 3D points by estimating a reliability mask over the 3D scene, as we describe in Section~\ref{sec:reliability}. Finally, we track in 3D, by re-locating the object within each timestep's 3D map, as described in Section~\ref{sec:tracking}. Figure~\ref{fig:overview} shows an overview of the training and testing setups.

\subsection{Neural 3D Mapping} \label{sec:neural_mapping}

Our model learns to map an RGB-D (RGB and depth) image to a 3D feature map of the scene in an end-to-end differentiable manner. The basic architecture is based on prior works \cite{commonsense,adam3d}, which proposed view prediction architectures with a 3D bottleneck. In our case, the 3D feature map is the \textit{output} of the model, rather than an intermediate representation.

Let $\M \in \mathbb{R}^{W \times H \times D \times C}$ denote the 3D feature map representation, where $W,H,D,C$ denote the width, height, depth and number of feature channels, respectively. The map corresponds to a large cuboid of world space, placed at some pose of interest (e.g., surrounding a target object). Every $(x,y,z)$ location in the 3D feature map $\M$ holds a $C$-length feature vector that describes the semantic and geometric content of the corresponding location of the world scene. To denote the feature map of timestep $i$, we write $\Mi$. We denote the function that maps RGB-D inputs to 3D feature maps as $f: (\I, \D) \mapsto \M$. To implement this function, we voxelize the inputs into a 3D grid, then pass this grid through a 3D convolutional network, and $L_2$-normalize the outputs. 


Tung et al.~\cite{commonsense} learned the parameters of $f$ by predicting RGB images of unseen viewpoints, and applying a regression loss; Harley et al.~\cite{adam3d} demonstrated that this can be outperformed by contrastive prediction objectives, in 2D and 3D. Here, we drop the view prediction task altogether, and focus entirely on a 3D correspondence objective: if a static point $(x,y,z)$ is observed in two views $i$ and $j$, the corresponding features $m_i = \Mi_{x,y,z}, m_j = \Mj_{x,y,z}$ should be similar to each other, and distinct from other features. We achieve this with a cross entropy loss \cite{sohn2016improved,oord2018representation,he2019momentum,chen2020simple}:
\begin{equation}
    \loss_{i,j} = -\log \frac{ \exp(  m_i^\top m_j / \tau)}{\sum_{k \neq i} \exp( m_i^\top m_k / \tau) },
\end{equation}
where $\tau$ is a temperature parameter, which we set to $0.07$, and the sum over $k \neq i$ iterates over non-corresponding features. Note that indexing correctly into the 3D scene maps to obtain the correspondence pair $m_i, m_j$ requires knowledge of the relative camera transformation across the input viewpoints; we encapsulate this registration and indexing in the notation $\M_{x,y,z}$. Following He et al.~\cite{he2019momentum}, we obtain a large pool of negative correspondences through the use of an offline dictionary, and stabilize training with a ``slow'' copy of the encoder, $f_\textrm{slow}$, which is learned via high-momentum updates from the main encoder parameters. 

Since the neural mapper (a) does not know a priori which voxels will be indexed for a loss, and (b) is fully convolutional, it learns to generate view-invariant features densely in its output space, even though the supervision is sparse. Furthermore, since (a) the model is encouraged to generate correspondable features invariant to viewpoint, and (b) varying viewpoints provide varying contextual support for 3D locations, the model learns to \textit{infer} corrrespondable features from limited context, which gives it robustness to partial occlusions. 

\subsection{Inferring static points for self-supervision in dynamic scenes} 
\label{sec:reliability}

The training objective in the previous subsection requires the location of a static point observed in two or more views. In a scenario where the data is made up entirely of static scenes, as can be achieved in simulation or in controlled environments, obtaining these static points is straightforward: any point on the surface of the scene will suffice, provided that it projects into at least two camera views. 

To make use of data where some parts are static and other parts are moving, we propose to simply discard data that appears to be moving. We achieve this by training a neural module to take the difference of two scene features $\Mi, \Mj$ as input, and output a ``reliability'' mask indicating a per-voxel confidence that the scene cube within the voxel is static: $ g : (\textbf{sg}(\Mi-\Mj)) \mapsto \R^{W,H,D}$, where $\textbf{sg}$ stops gradients from flowing from $g$ into the function $f$ which produces $\Mi, \Mj$. We implement $g$ as a per-voxel classifier, with 2-layer fully-connected network applied fully convolutionally. We do not assume to have true labels of moving/static voxels, so we generate synthetic labels using static data: given two maps of the \textit{same} scene $\Mi, \Mj$, we generate positive-label inputs with $\Mi - \Mj$ (as normal), and generate negative-label inputs with $\Mi - \textrm{shuffle}(\Mj)$, where the shuffle operation ruins the correspondence between the two tensors. After this training, we deploy this network on pairs of frames from dynamic scenes, and use it to select high-confidence static data to further train the encoder $f$. 

Our training procedure is then: (1) learn the encoder $f$ on static data; (2) learn the reliability function $g$; (3) in dynamic data, finetune $f$ on data selected by $g$. Steps 2-3 can be repeated a number of times. In practice we find that results do not change substantially after the first pass. 

\subsection{Tracking via point correspondences} \label{sec:tracking}

Using the learned neural 3D mapper, we track an object of interest over long temporal horizons by re-locating it in a map produced at each time step. Specifically, we re-locate each voxel of the object, and use these new locations to form a motion field. We convert these voxel motions into an estimate for the entire object by fitting a rigid transformation to the correspondences, via RANSAC. 

We assume we are given the target object's 3D box on the zeroth frame of a test video. Using the zeroth RGB-D input, we generate a 3D scene map centered on the object. We then convert the 3D box into a set of coordinates $X_0, Y_0, Z_0$ which index into the map. Let $m_i = \Mz_{x_i \in X_0, y_i \in Y_0, z_i \in Z_0}$ denote a voxel feature that belongs to the object. On any subsequent frame $t$, our goal is to locate the new coordinate of this feature, denoted $x_i^t, y_i^t, z_i^t$. We do so via a soft spatial argmax, using the learned feature space to provide correspondence confidences: 
\begin{equation}
    (x_i^t, y_i^t, z_i^t) = \sum_{x \in X, y \in Y, z \in Z}  \left( \frac{ \exp(m_i^\top \Mt_{x,y,z})}{\sum_{x_2 \in X, y_2 \in Y, z_2 \in Z} \exp( m_i^\top \Mt_{x_2,y_2,z_2} )}(x,y,z) \right),
    \label{eq:reloc}
\end{equation}
where $X, Y, Z$ denote the set of coordinates in the search region. We then compute the motion of the voxel as $(\delta_{x_i}^t, \delta_{y_i}^t, \delta_{z_i}^t) = (x_i^t, y_i^t, z_i^t) - (x_i, y_i, z_i)$. After computing the motion of every object voxel in this way, we use RANSAC to find a rigid transformation that explains the majority of the correspondences. 
We apply this rigid transform to the input box, yielding the object's location in the next frame. 

Vondrick et al.~\cite{vondrick2018tracking} computed a similar attention map during tracking (in 2D), but did not compute its soft argmax, nor explain the object's motion with a single transformation, but rather propagated a ``soft mask'' to the target frame, which is liable to grow or shrink. Our method takes advantage of the fact that all coordinates are 3D, and makes the assumption that the objects are rigid, and propagates a fixed-size box from frame to frame. 

We empirically found that it is critical to constrain the search region of the tracker in 3D. In particular, on each time step we create a search region centered on the object's last known position. The search region is $16m \times 2m \times 16m$ large, which is half of the typical full-scene resolution. This serves three purposes. The first is: it limits the number of spurious correspondences that the model can make, since it puts a cap on the metric range of the correspondence field, and thereby reduces errors. Second: it ``re-centers'' the model's field of view onto the target, which allows the model to incorporate maximal contextual information surrounding the target. Even if the bounds were sufficiently narrow to reduce spurious correspondences, an object at the \textit{edge} of the field of view will have less-informed features than an object at the middle, due to the model's convolutional architecture. The third reason is computational: even 2D works \cite{vondrick2018tracking} struggle with the computational expense of the large matrix multiplications involved in this type of soft attention, and in 3D the expense is higher. Searching locally instead of globally makes Eq.~\ref{eq:reloc} tractable.  



\begin{figure}[t!]
 \centering
 \includegraphics[width=1.0\linewidth]{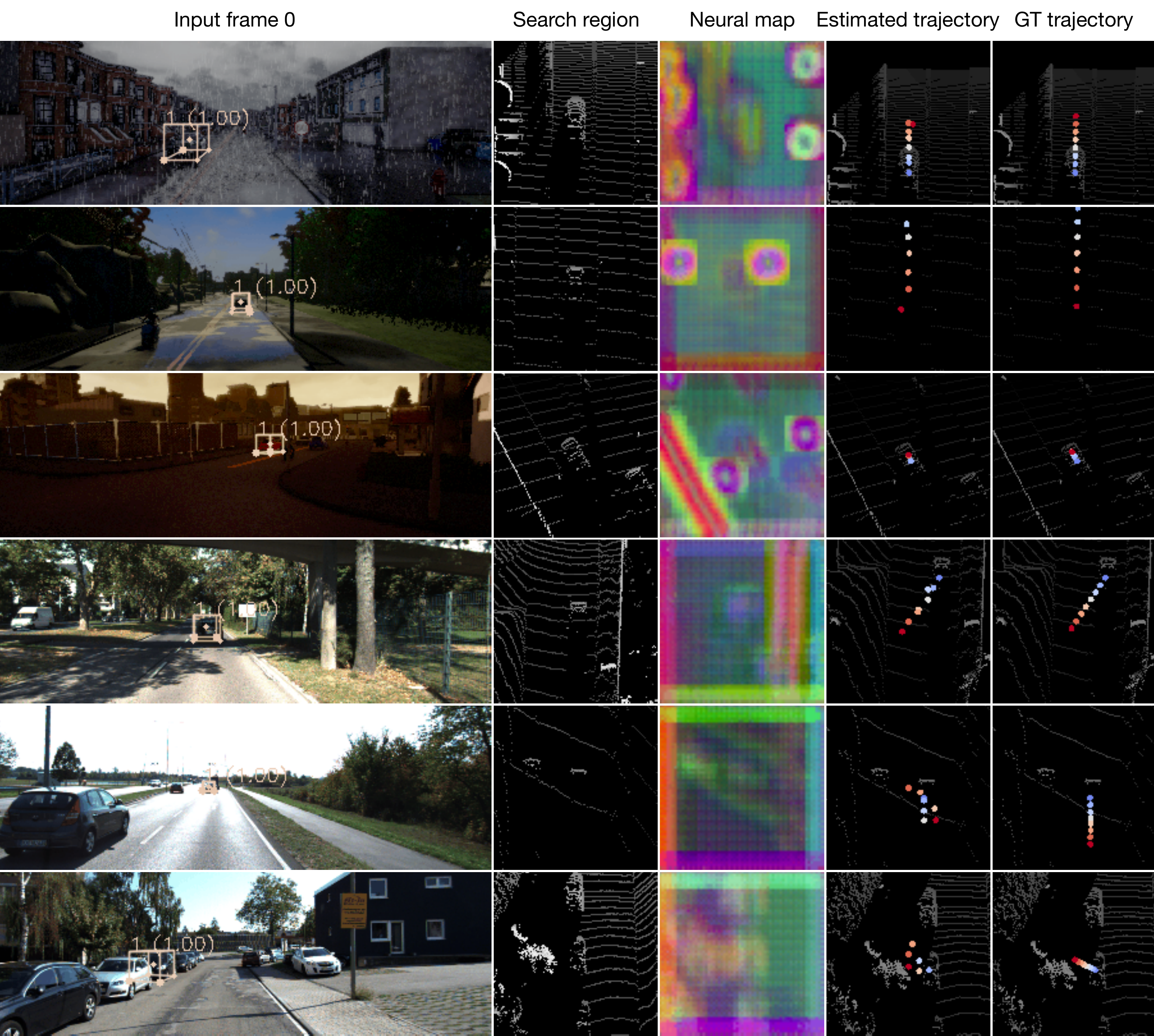}
 \caption{\textbf{Visualization of tracking inputs, inferred 3D scene features, and tracking outputs.} Given the input frame on the left, along with a 3D box specifying the object to track, the model (1) creates a search region for the object centered on the object's position (bird's eye view of occupancy shown), (2) encodes this region into a 3D neural map (bird's eye view of PCA compression shown), and (3) finds correspondences for the object's features. The top three rows show CARLA results; the bottom three rows show KITTI results. 
 }
 \label{fig:tracking3d_qual}
\end{figure}

\section{Experiments}\label{sec:exp}

\begin{figure}[t]
 \centering
 \includegraphics[width=1.0\linewidth]{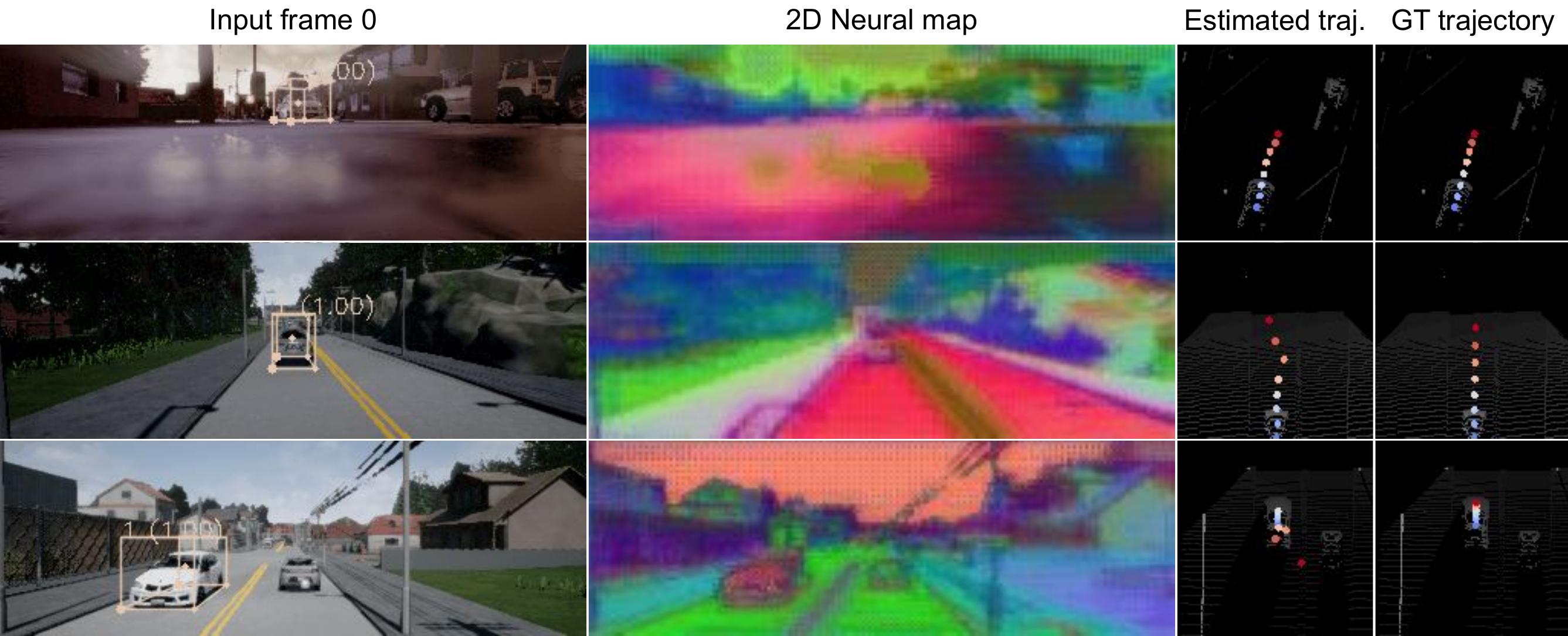}
 \caption{\textbf{Visualization of tracking inputs, inferred 2D scene features, 
 and tracking outputs, for our adaptation of Dense Object Nets \cite{florence2018dense}.} This model operates in the same way as our full 3D one, but creates features in 2D, and unprojects them to a sparse 3D grid for tracking. 
 }
 \label{fig:tracking2d_qual}
\end{figure}

We test our model in the following two datasets: 
\begin{enumerate} 
\item \textbf{Synthetic RGB-D videos of urban scenes rendered in the CARLA simulator \cite{Dosovitskiy17}}. CARLA is an open-source photorealistic simulator of urban driving scenes. It permits moving the camera to any desired viewpoint in the scene. 
We obtain data from the simulator as follows. We begin by generating 10000 autopilot episodes of 16 frames each, at 10 FPS. 
We define 18 viewpoints along a 40m-radius hemisphere anchored to the ego-car (i.e., it moves with the car). 
In each episode, we sample 6 random viewpoints from the 18 and randomly perturb their pose, and then capture each timestep of the episode from these 6 viewpoints. We discard episodes that do not have an object in-bounds for the duration of the episode. 

We treat the Town1 data as the ``training'' set, and the Town2 data as the ``test'' set, so there is no overlap between the train and test sets. This yields 4313 training videos, and 2124 test videos. 

\item \textbf{Real RGB-D videos of urban scenes, from the KITTI dataset \cite{Geiger2013IJRR}.} This data was collected with a sensor platform mounted on a moving vehicle, with a human driver navigating through a variety of areas in Germany. We use the ``left'' color camera, and LiDAR sweeps synced to the RGB images. 

For training, we use the ``odometry'' subset of KITTI; it includes egomotion information accurate to within 10cm. 
The odometry data includes ten sequences, totalling 23201 frames. 

We test our model in the validation set of the ``tracking'' subset of KITTI, which has twenty labelled sequences, totalling 8008 frames. For supervised baselines, we split this data into 12 training sequences and 8 test sequences. For evaluation, we create 8-frame subsequences of this data, in which a target object has a valid label for all eight frames. This subsequencing is necessary since objects are only labelled when they are within image bounds. The egomotion information in the ``tracking'' data is only approximate. 
\end{enumerate}
We evaluate our model on its ability to track objects in 3D. On the zeroth frame, we receive the 3D box of an object to track. On each subsequent frame, we estimate the object's new 3D box, and measure the intersection over union (IOU) of the estimated box with the ground truth box. We report IOUs as a function of timesteps. 
 
\subsection{Baselines} 

We evaluate the following baselines. We provide additional implementation details for each baseline (and for our own model) in the supplementary file. 

\begin{itemize}
\item \textbf{Unsupervised 3D flow \cite{adam3d}.} This model uses an unsupervised architecture similar to ours, but with a 2-stage training procedure, in which features are learned first from static scenes and frozen, then a 3D flow module is learned over these features in dynamic scenes. We extend this into a 3D tracker by ``chaining'' the flows across time, and by converting the trajectories into rigid motions via RANSAC. 


\item \textbf{2.5D dense object nets \cite{florence2018dense}.} This model learns to map input images into dense 2D feature maps, and uses a contrastive objective at known correspondences across views. We train this model using static points for correspondence labels (like our own model). 
We extend this model into a 3D tracker by ``unprojecting'' the learned embeddings into sparse 3D scene maps, then applying the \textit{same tracking pipeline} as our own model. 


\item \textbf{2.5D tracking by colorization \cite{vondrick2018tracking,mast}.} This model learns to map input images into dense 2D feature maps, using an an RGB reconstruction objective. The model trains as follows: given two RGB frames, the model computes a feature map for each frame; for each pixel of the first frame's feature map, we compute that feature's similarity with all features of the second frame, and then use that similarity matrix to take a weighted combination of the second frame's colors; this color combination at every pixel is used as the reconstruction of the first frame, which yields an error signal for learning. 
We extend this model into a 3D tracker in the same way that we extended the ``dense object nets'' baseline. 

\item \textbf{3D neural mapping with random features.} This model is equivalent to our proposed model but with randomly-initialized network parameters. 
This model may be expected to perform at better-than-chance levels due to the power of random features \cite{rahimi2008random} and due to the domain knowledge encoded in the architecture design. 

\item \textbf{3D fully convolutional siamese tracker (supervised) \cite{bertinetto2016fully}.} This is a straightforward 3D upgrade of a fully convolutional 2D siamese tracker, which uses the object's feature map as a cross correlation template, and tracks the object by taking an argmax of the correlation heatmap at each step. It is necessary to supervise this model with ground-truth box trajectories. 
We also evaluate a ``cosine windowing'' variant of this model, which suppresses correlations far from the search region's centroid \cite{bertinetto2016fully}.  

\item \textbf{3D siamese tracker with random features.} This model is equivalent to the 3D siamese supervised tracker, but with randomly-initialized network parameters. Similar to the random version of 3D neural mapping, this model measures whether random features and the implicit biases are sufficient to track in this domain.

\item \textbf{Zero motion.} This baseline simply uses the input box as its final estimate for every time step. This baseline provides a measure for how quickly objects tend to move in the data, and serves as a lower bound for performance. 
\end{itemize}
All of these are fairly simple trackers. A more sophisticated approach might incrementally update the object template \cite{matthews2004template}, but we leave that for future work. 

\paragraph{Ablations.} We compare our own model against the following ablated versions. First, we consider a model without search regions, which attempts to find correspondences for each object point in the \textit{entire} 3D map at test time. This model is far more computationally expensive, since it requires taking the dot product of each object feature with the entire scene. Second, we consider a similar ``no search region'' model but at half resolution, which brings the computation into the range of the proposed model. This ablation is intended to reveal the effect of resolution on accuracy. Third, we omit the ``static point selection'' (via the function $g$). This is intended to evaluate how correspondence errors caused by moving objects (violating the static scene assumption) can weaken the model.

\subsection{Quantitative results}

\begin{figure}[t!]
 \centering
 \includegraphics[width=1.0\linewidth]{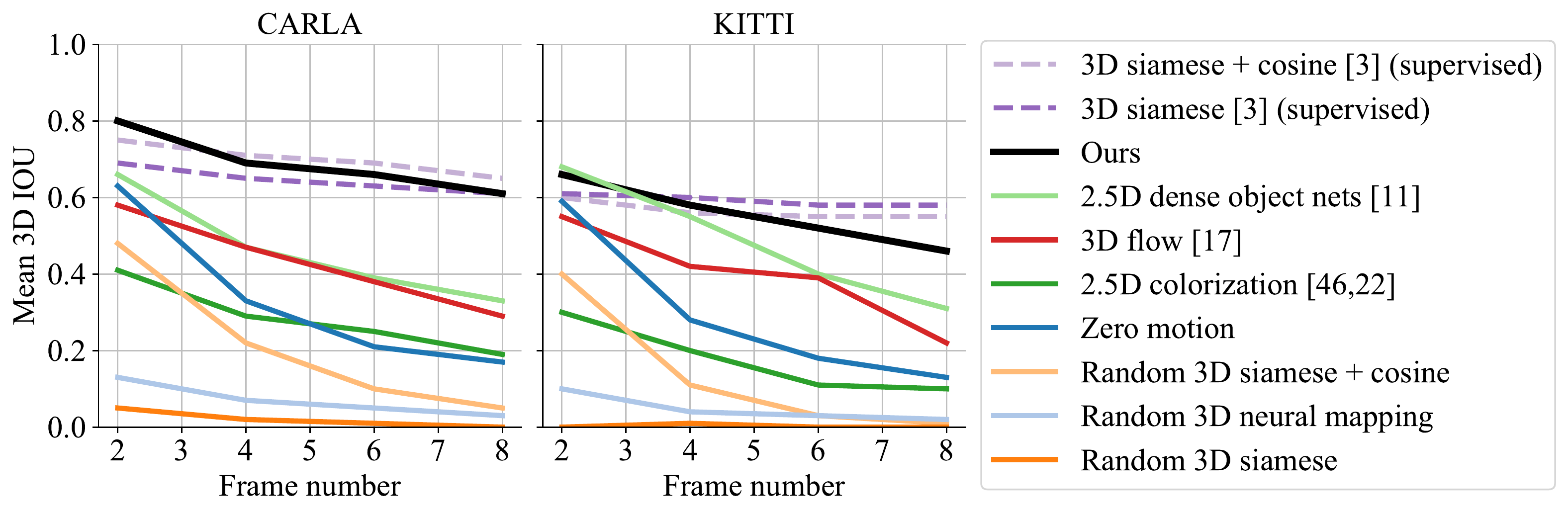}
 \caption{ 
 \textbf{Single-object tracking accuracy, in mean IOU across time, in CARLA (left) and KITTI (right).} Methods are sorted in the legend in decreasing order of mean IOU. }
 \label{fig:all_ious}
\end{figure}

We evaluate the mean 3D IOU of our trackers over time. Figure~\ref{fig:all_ious}-left 
shows the results of this evaluation in CARLA.
As might be expected, the supervised 3D trackers perform best, and cosine windowing improves results. 



Our model outperforms all other unsupervised models, and nearly matches the supervised performance. 
The 2.5D dense object net performs well also, 
but its accuracy is likely hurt by the fact that it is limited exclusively to points observed in the depth map. Our own model, in contrast, can match against both observed and unobserved (i.e., hallucinated or inpainted) 3D scene features. The colorization model performs under the 2.5D dense object net approach, likely because this model only indirectly encourages correspondence via the colorization task, and therefore is a weaker supervision than the multi-view correspondence objectives used in the other methods. 

Random features perform worse than the zero-motion baseline, both with a neural mapping architecture and a siamese tracking architecture. Inspecting the results qualitatively, it appears that these models quickly propagate the 3D box off of the object and onto other scene elements. This suggests that random features and the domain knowledge encoded in these architectures are not enough to yield 3D trackers in this data. 


We perform the same evaluation in KITTI, and show the results in Figure~\ref{fig:all_ious}-right.  
On this benchmark, accuracies are lower for all models, indicating that the task here is more challenging. This is likely related to the fact that (1) the egomotion is imperfect in this data, and (2) the tracking targets are frequently farther away than they are in CARLA. Nonetheless, the ranking of methods is the same, with our 3D neural mapping model performing best. One difference is that cosine windowing actually worsens siamese tracking results in KITTI. This is likely related to the fact that egomotion stabilization is imperfect in KITTI: a zero-motion prior is only helpful in frames where the target is stationary \textit{and} camera motion is perfectly accounted for; otherwise it is detrimental. 

We additionally split the evaluation on stationary vs. moving objects in CARLA. We use a threshold of $1m$ total distance (in world coordinates) across the 8-frame trajectories to split these categories. At the last timestep, the mean 3D IOU for all objects together it is $0.61$ (as shown in Figure~\ref{fig:all_ious}-left); for static objects only, the value is $0.64$; for moving objects only, it is $0.56$. This suggests that the model tracks stationary objects more accurately than moving objects, likely because their appearance changes less with respect to the camera and the background. 

Finally, we evaluate the top two models in CARLA using the standard 2D tracking metrics, multi-object tracking accuracy (MOTA) and multi-object tracking precision (MOTP) \cite{bernardin2006multiple}, though we note that our task only has one target object per video. We find that the 3D siamese + cosine model (supervised) achieves a MOTA of $0.9578$ and MOTP of $0.7407$, while our model achieves MOTA $0.9125$ and MOTP $0.8384$. This suggests that the supervised tracker makes fewer misses, but our method delivers slightly better precision.

\subsection{Qualitative results} 

We visualize our tracking results, along with inputs and colorized visualizations of our neural 3D scene maps, in Figure~\ref{fig:tracking3d_qual}. To visualize the neural 3D maps, we take a mean along the vertical axis of the grid (yielding a ``bird's eye view'' 2D grid), compress the deep features in each grid cell to 3 channels via principal component analysis, normalize, and treat these values as RGB intensities. 
Comparing the 3D features learned in CARLA vs those learned in KITTI reveals a very obvious difference: the KITTI features appear blurred and imprecise in comparison with the CARLA features. This is likely due to the imperfect egomotion information, which leads to slightly inaccurate correspondence data at training time (i.e., occasional failures by the static point selector $g$).

In Figure~\ref{fig:tracking2d_qual}, we visualize the features learned by Dense Object Nets in this data. From the PCA colorization it appears that objects are typically colored differently from their surroundings, which is encouraging, but the embeddings are not as clear as those in the original work~\cite{florence2018dense}, likely because this domain does not have the benefit of background image subtraction and object-centric losses. 

In the supplementary file, we include video visualizations of the learned features and 3D tracking results. 

\subsection{Ablations}

We evaluate ablated versions of our model, to reveal the effect of (1) search regions, (2) resolution, and (3) static point selection in dynamic scenes. Results are summarized in  Table~\ref{tab:ablations}. 

Without search regions, the accuracy of our model drops by 20 IOU points, which is a strong impact. We believe this drop in performance comes from the fact that search regions take advantage of 3D scene constancy, by reducing spurious correspondences in far-away regions of the scene. 

Resolution seems to have a strong effect as well: halving the resolution of the wide-search model  reduces its performance by 15 IOU points. This may be related to the fact that fewer points are then available for RANSAC to find a robust estimate of the object's rigid motion. 

Since static point selection is only relevant in data with moving objects, we perform this experiment in KITTI (as opposed to CARLA, where the training domain is all static by design).  The results show that performance degrades substantially without the static point selection. This result is to be expected, since this ablation causes erroneous correspondences to enter the training objective, and thereby weakens the utility of the self-supervision.

\setlength{\tabcolsep}{4pt}
\begin{table}
\begin{center}
\caption{Ablations of our model.}
\label{tab:ablations}
\begin{tabular}{lc}
\hline\noalign{\smallskip}
Method & Mean IOU \\
\noalign{\smallskip}
\hline
\noalign{\smallskip}
Ours in CARLA  & 0.61\\
\ldots without search regions & 0.40\\
\ldots without search regions, at half resolution & 0.25\\
Ours in KITTI  & 0.46\\
\ldots without static point selection & 0.39\\
\hline
\end{tabular}
\end{center}
\end{table}
\setlength{\tabcolsep}{1.4pt}

\subsection{Limitations}\label{sec:limitations}

The proposed model has three important limitations. 
First, 
our work assumes access to RGB-D data with accurate egomotion data at training time, with a wide variety of viewpoints. This is easy to obtain in simulators, but real-world data of this sort typically lies along straight trajectories (as it does in KITTI), which limits the richness of the data. 
Second, our model architecture requires a lot of GPU memory, due to its third spatial dimension. This severely limits either the resolution or the metric span of the latent map $\M$. 
On 12G Titan X GPUs we encode a space sized $32m \times 4m \times 32m$ at a resolution of $128 \times 32 \times 128$, with a batch size of 4; iteration time is ~0.2s/iter. 
Sparsifying our feature grid, or using 
points instead of voxels, 
are clear areas for future work. 
Third, our test-time tracking algorithm makes two strong assumptions: (1) a tight box is provided in the zeroth frame, and (2) the object is rigid. For non-rigid objects, merely propagating the box with the RANSAC solution would be insufficient, but the voxel-based correspondences might still be helpful. 

\section{Conclusion} \label{sec:conclusion}
We propose a model which learns to track objects in dynamic scenes just from observing static scenes. 
We show that a multi-view contrastive loss allows us to learn rich visual representations that are correspondable not only across views, but across time.
We demonstrate the robustness of the learned representation by benchmarking the learned features on a tracking task in real and simulated data. Our approach outperforms prior unsupervised 2D and 2.5D trackers, and approaches the accuracy of supervised trackers. 
Our 3D representation benefits from denser correspondence fields than 2.5D methods, and is invariant to the artifacts of camera projection, such as apparent scale changes of objects.
Our approach opens new avenues for learning trackers in arbitrary environments, without requiring explicit tracking supervision: if we can obtain an accurate pointcloud reconstruction of an environment, then we can learn a tracker for that environment too. 

\section*{Acknowledgements}
This material is based upon work funded and supported by the Department of Defense under Contract No. FA8702-15-D-0002 with Carnegie Mellon University for the operation of the Software Engineering Institute, a federally funded research and development center. We also acknowledge the support of the Natural Sciences and Engineering Research Council of Canada (NSERC), AiDTR, the DARPA Machine Common Sense program, and the AWS Cloud Credits for Research program.

\clearpage

\title{Tracking Emerges by Looking Around Static Scenes, with Neural 3D Mapping -- Supplementary Material}

\author{}
\institute{}

\maketitle

\section{Implementation details}

\subsection{Neural 3D mapping}

Our 3D convolutional network $f$ has the following architecture. The input is a 4-channel 3D grid shaped $128 \times 32 \times 128$. The four channels are RGB and occupancy. ``Unoccupied'' voxels have zeros in all channels. This input is passed through a 3D convolutional network, which is an encoder-decoder with skip connections. There are three encoder layers: each have stride-2 3D convolutions, with kernel size $4 \times 4 \times 4$, and the output channel dimensions are $64, 128, 192$. There are three decoder layers: the first two have stride-2 transposed convolutions with stride 2 and $4 \times 4 \times 4$ kernels with $256$ output channels; after each of these we concatenate the same-resolution layer from the encoder half; the last layer has a $1 \times 1 \times 1$ kernel and $64$ channels. 

At training time $f$ encodes a metric space sized $32 \times 4 \times 32$ meters. At test time we ``zoom in'' on the area of interest (i.e., the search region), and encode a space sized $16 \times 4 \times 16$ meters.

We train with the Adam optimizer for $200,000$ iterations at a learning rate of $1e-4$. 

\subsection{Baselines}

\begin{itemize}
\item \textbf{Unsupervised 3D flow \cite{adam3d}.} 
The original work by Harley et al.~\cite{adam3d} only estimated flow for pairs of frames. We extend this into a 3D tracker by deploying the flow module in an object-centric manner, and by ``chaining'' the flows together across time. We effectively compute the flow from the first frame \textit{directly} to every other frame in the sequence, but with an alignment step designed to keep the object within the field of view of the flow module. Specifically, given the object location in the first frame, we extract a volume of flow vectors in the object region, and run RANSAC to find a rigid transformation that explains the majority of the flow field. We apply this rigid transform to the box, yielding the object's location in the next frame. Then, we back-warp the next frame according to the box transformation, to re-center the voxel grid onto the object. Estimating the flow between the original frame and the newly backwarped frame yields the \textit{residual} flow field; adding this residual to the original flow provides the new cumulative motion of the object. We repeat these steps across the length of the input video, back-warping according to the cumulative flow and estimating the residual. 

\item \textbf{2.5D dense object nets \cite{florence2018dense}.} While the original work used a margin loss for contrastive learning, we update this to a cross entropy loss with a large offline dictionary and a coupled ``slow'' encoder, consistent with state-of-the-art contrastive learning \cite{he2019momentum}. We extend this model into a 3D tracker as follows. First, we use the available depth data to ``unproject'' the learned embeddings into sparse 3D scene maps. Then, we use the same tracking pipeline as our own model (with search regions and soft argmaxes and RANSAC). The differences between this model and our own are (1) it learns a 2D CNN instead of a 3D one, and (2) its 3D output is constrained to the locations observed in the depth map, instead of producing features densely across the 3D grid. 

\item \textbf{2.5D tracking by colorization \cite{vondrick2018tracking,mast}.} We use the latest iteration of this method \cite{mast}, which uses the LAB colorspace at input and output, and color dropout at training time; this outperforms the quantized-color cross entropy loss of the original work. 
We extend this model into a 3D tracker in the same way that we extended the ``dense object nets'' baseline. Therefore, the only difference between the two 2.5D methods is in the training: this method can train on arbitrary data but has an objective that only indirectly encourages correspondence (through an RGB reconstruction loss), while ``2.5D dense object nets'' requires static scenes and directly encourages correspondences (through a contrastive loss).

\item \textbf{3D neural mapping with random features.} We use the same inputs, architecture, outputs, and the same hyperparameters for resolution and search regions, but do not train the parameters. 

\item \textbf{3D fully convolutional siamese tracker (supervised) \cite{bertinetto2016fully}.} We use the same resolutions and feature encoders for this model as we do for our main model. The cosine window adds a zero-velocity bias into the model, and makes it less likely for the model's outputs to make large jumps \cite{bertinetto2016fully}.
\end{itemize}

\section{Detailed results}

In Tables~\ref{tab:tracking_carla} and~\ref{tab:tracking_kitti}, we provide the numerical values of the data plotted in the main paper. The notation IOU@N denotes 3D intersection over union at the Nth frame of the video. Note that IOU@0 $= 1.0$ in this task, since tracking is initialized with a ground-truth box in the zeroth frame. 

\setlength{\tabcolsep}{4pt}
\begin{table}
\begin{center}
\caption{Single-object tracking accuracy, in mean IOU across time, in CARLA.}
\label{tab:tracking_carla}
\begin{tabular}{lcccc}
\hline\noalign{\smallskip}
Method & IOU@2 & IOU@4 & IOU@6 & IOU@8 \\
\noalign{\smallskip}
\hline
\noalign{\smallskip}
Zero motion  & 0.63 & 0.33 & 0.21 & 0.17\\
Random 3D neural mapping  & 0.13 & 0.07 & 0.05 & 0.03\\
Random 3D siamese  & 0.05 & 0.02 & 0.01 & 0.00\\
Random 3D siamese + cosine  & 0.48 & 0.22 & 0.10 & 0.05\\
2.5D colorization~\cite{vondrick2018tracking,mast}  & 0.41 & 0.29 & 0.25 & 0.19\\
2.5D dense object nets~\cite{florence2018dense}  & 0.66 & 0.47 & 0.39 & 0.33\\
3D flow~\cite{adam3d}  & 0.58 & 0.47 & 0.38 & 0.29\\
Ours  & 0.80 & 0.69 & 0.66 & 0.61\\
\hline 
3D siamese~\cite{bertinetto2016fully} (supervised)  & 0.69 & 0.65 & 0.63 & 0.61\\
3D siamese + cosine~\cite{bertinetto2016fully} (supervised)  & 0.75 & 0.71 & 0.69 & 0.65\\
\hline
\end{tabular}
\end{center}
\end{table}
\setlength{\tabcolsep}{1.4pt}


\setlength{\tabcolsep}{4pt}
\begin{table}
\begin{center}
\caption{Single-object tracking accuracy, in mean IOU across time, in KITTI.}
\label{tab:tracking_kitti}
\begin{tabular}{lcccc}
\hline\noalign{\smallskip}
Method & IOU@2 & IOU@4 & IOU@6 & IOU@8 \\
\noalign{\smallskip}
\hline
\noalign{\smallskip}
Zero motion  & 0.59 & 0.28 & 0.18 & 0.13\\
Random 3D neural mapping  & 0.10 & 0.04 & 0.03 & 0.02\\
Random 3D siamese & 0.00 & 0.01 & 0.00 & 0.00\\
Random 3D siamese + cosine & 0.40 & 0.11 & 0.03 & 0.01\\
2.5D colorization~\cite{vondrick2018tracking,mast}  & 0.30 & 0.20 & 0.11 & 0.10\\
2.5D dense object nets~\cite{florence2018dense}  & 0.68 & 0.55 & 0.40 & 0.31\\
3D flow~\cite{adam3d}  & 0.55 & 0.42 & 0.39 & 0.22\\
Ours  & 0.66 & 0.58 & 0.52 & 0.46\\
\hline 
3D siamese~\cite{bertinetto2016fully} (supervised)  & 0.61 & 0.60 & 0.58 & 0.58\\
3D siamese + cosine~\cite{bertinetto2016fully} (supervised) & 0.60 & 0.56 & 0.55 & 0.55\\
\hline
\end{tabular}
\end{center}
\end{table}
\setlength{\tabcolsep}{1.4pt}

\end{document}